\documentclass[graybox]{svmult}

\usepackage{mathptmx}       
\usepackage{helvet}         
\usepackage{courier}        
\usepackage{type1cm}        

\usepackage{makeidx}         
\usepackage{graphicx}        
\usepackage{multicol}        

\usepackage{verbatim}
\usepackage{amsmath}
\usepackage{amssymb}
\usepackage{subfig}
\usepackage{url}

\begin{document}

\title*{Nuclear Environments Inspection with Micro Aerial Vehicles: Algorithms and Experiments}
\author{Dinesh Thakur$^*$$^{1}$, Giuseppe Loianno$^*$$^{2}$, Wenxin Liu$^{1}$, and Vijay Kumar$^{1}$}

\institute{$^{1}$The authors are with the GRASP Lab, University of Pennsylvania, 3330 Walnut Street, 19104 Philadelphia, USA.{\tt\small \{tdinesh, wenxinl, kumar\}@seas.upenn.edu.}\\
$^{2}$The author is with the New York University, Tandon School of Engineering, 6 MetroTech Center, 11201 Brooklyn NY, USA. {\tt\small loiannog@nyu.edu.}\\
$^*$ These authors contributed equally}

\maketitle

\abstract{In this work, we address the estimation, planning, control and mapping problems to allow a small quadrotor to autonomously inspect the interior of hazardous damaged nuclear sites. These algorithms run onboard on a computationally limited CPU. We investigate the effect of varying illumination on the system performance. To the best of our knowledge, this is the first fully autonomous system of this size and scale applied to inspect the interior of a full scale mock-up of a  Primary Containment Vessel (PCV). The proposed solution opens up new ways to inspect nuclear reactors and to support nuclear decommissioning, which is well known to be a dangerous, long and tedious process. Experimental results with varying illumination conditions show the ability to navigate a full scale mock-up PCV pedestal and create a map of the environment, while concurrently avoiding obstacles.}
\keywords{Inspection, Aerial Robotics}

\section{Introduction}

\begin{figure*}[t]
  \centering
  \includegraphics[height=3.5cm, trim={1cm 0 0 2cm},clip]{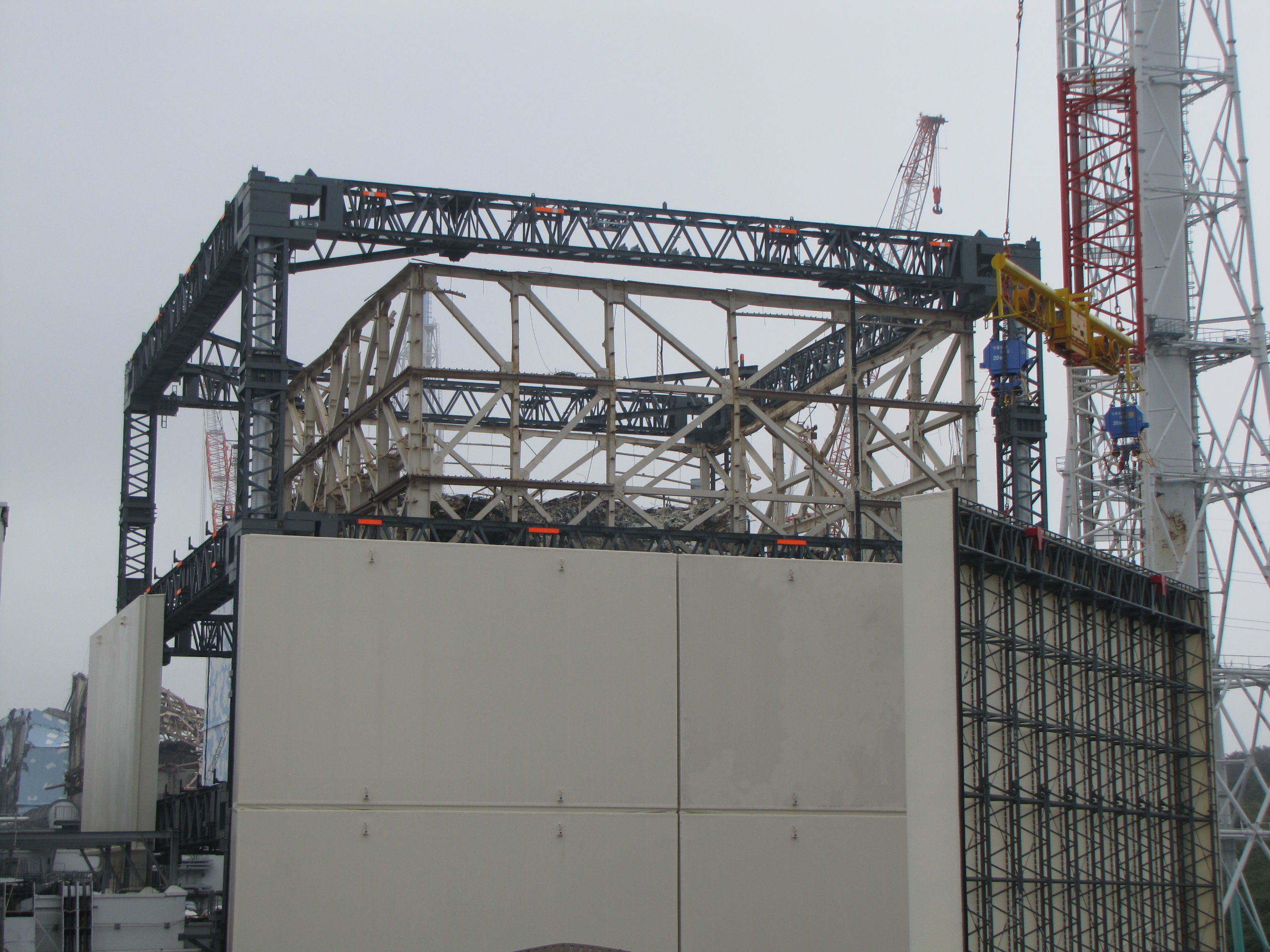}
  \hspace{0.1cm}
  \includegraphics[height=3.5cm, trim={4cm 1cm 2cm 0},clip]{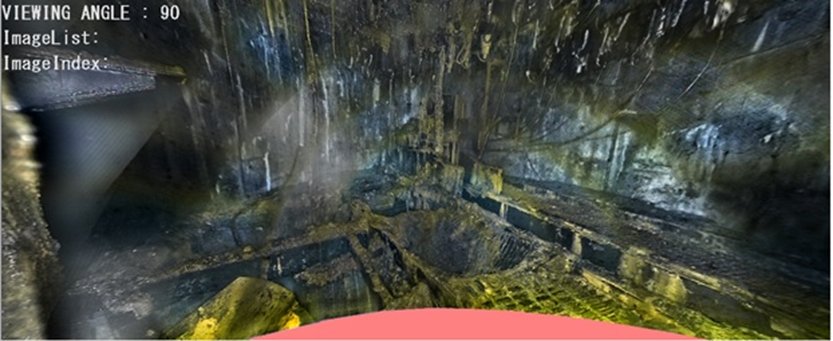}
  \caption{Left - the exterior of one of the damaged units at Fukushima Daiichi~\cite{decomissioning_tepco}, Right - processed image of the pedestal area inside of Unit 2 collected after the meltdown~\cite{decomissioning_tepco}.}
  \label{fig:decomissioning}
\end{figure*}

\begin{figure*}[t]
  \centering
  \includegraphics[height=3.55cm]{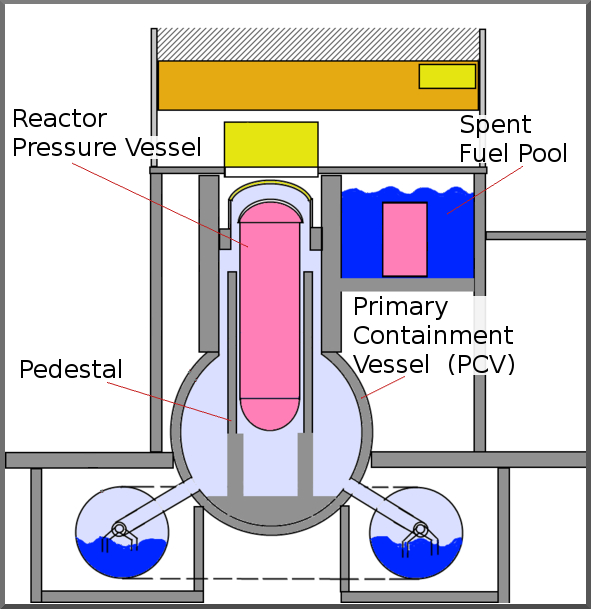}
  \hspace{0.1in}
  \includegraphics[height=3.5cm,trim={0 0 0 2cm},clip]{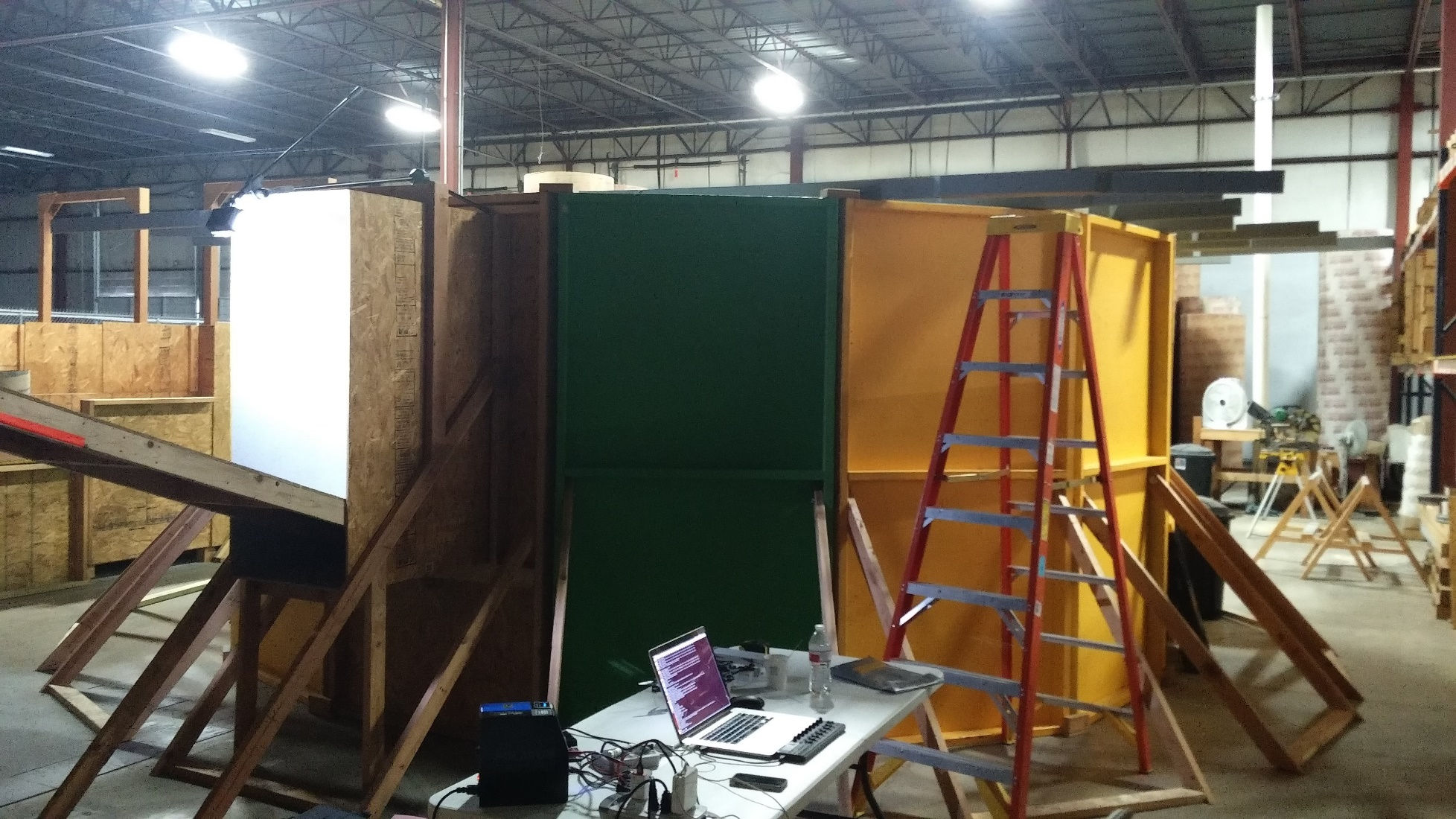}
  \caption{Cross-sectional view of a BWR containment (left)~\cite{bwr_containment}. Mock-up of the PCV pedestal and control rod drive (CRD) ramp (right).}
  \label{fig:mock_up}
\end{figure*}

Nuclear site decommissioning is a complex and tedious process, which includes clean-up of radioactive materials and progressive dismantling of the site as shown in Fig.~\ref{fig:decomissioning}. Because of the finite life of a reactor, the decommissioning is an essential step in a nuclear power plant. This process is also critical if there is an accident, as was the case in the Fukushima Daiichi Nuclear Power Plant in 2011~\cite{website_world_nuclear}. For planning the decontamination and decommission, surveys of the inside of the containment vessel are crucial. Fig.~\ref{fig:mock_up} shows a simplified cross-section of a multi-story containment for a boiling water reactor (BWR). The primary goal of this work is to develop a fully autonomous system that is capable of inspecting inside damaged sites such as ones seen in Fig.~\ref{fig:decomissioning}. We consider an example problem scenario of visually inspecting inside the Primary Containment Vessel (PCV) of a damaged nuclear power plant unit.

Micro Aerial Vehicles (MAVs), equipped with on-board sensors, are ideal platforms for autonomous inspection of cluttered, confined, and hazardous environments. The use of aerial platforms in nuclear settings poses several challenges such as the need to navigate in damaged/unknown environments, without GPS, under low illumination, without communication link to human-pilot or user interface, etc. Furthermore, a damaged nuclear site generally poses constraints on access (entry hatches can be less than 0.1-0.3m in diameter) and the operating conditions can be adversely affected by condensation and fog.

A survey of various autonomous systems deployed in inspection of damaged areas is provided in~\cite{6386301_rescue_robots, Murphy2016}. A combination of ground and air vehicle is used in~\cite{Michael_JFR} to inspect a building after an earthquake. Autonomous inspection of penstocks and tunnels using aerial vehicles is done in~\cite{Ozaslan2016}.

Focusing on nuclear environment, wall climbing robots have been popular for inspecting undamaged steam generators at nuclear sites \cite{wall_climb_Li2017}. The authors in~\cite{aerial_mapping_01431161.2016.1252474} provide an overview of aerial robots for radiation mapping, while~\cite{drone_radio_6172936} develops a drone equipped with sensors for nuclear risk characterization. Inspection of outdoor environments is primarily done with the aid of GPS. In the case of GPS-denied environments, remote teleoperation is used~\cite{teleop_s17102234}. In~\cite{daiichi_01439911211249715} a robot is used in nuclear sites for damage assessment. Unmanned construction equipment and robots were used for surveillance work and cleaning up rubble outside buildings~\cite{daiichi_6106792}. Finally,~\cite{robots_tepco} provides various robots that have been deployed in nuclear settings.

All the previous solutions were remotely teleoperated without considering the autonomy component required by the complex and remote operations in a nuclear (eventually damaged) scenario. Moreover, the physical size of the robots range from 1 m to 10 m. These systems are big with weights usually ranging from 2 kg to more than 1000 kg and can carry bulky sensors. Our problem of navigating inside inside a PCV is very different. Previous attempts to visually inspect the reactor Unit 1 with ground robots have been unable to complete the missions as they got stuck in confined environments of the damaged sites.

This work presents multiple contributions. First, we develop a fully autonomous system with state estimation, control and mapping modules for PCV inspection task concurrently running onboard a custom designed 0.16 m platform. Second, using the onboard map created during the navigation, the vehicle is able to automatically replan its path to avoid obstacles. Finally, we conduct a set of preliminary studies and experiments in different conditions in a mock-up representing a PCV pedestal at Southwest Research Institute (SwRI). To the best of our knowledge, this is the first fully autonomous system of this size and scale applied to  inspect the interior of a full scale mockup PCV for supporting nuclear decommissioning. Although we motivate by specific example of inspecting inside a PCV where the entry point is of order of 0.1 to 0.2 m, this system can be used in autonomous inspection of any damaged/tight regions.

\section{Technical Approach}
\label{sec:2}

\begin{figure*}[t]
  \centering
  \includegraphics[width=5cm]{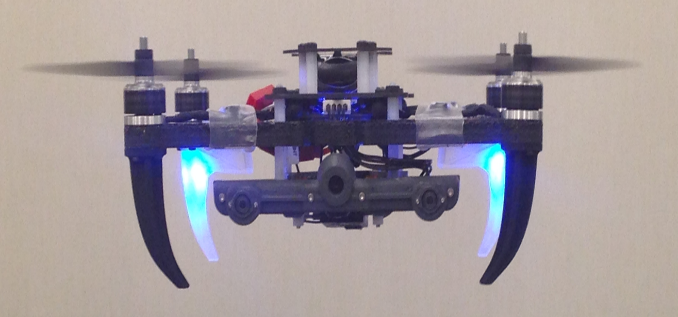}
  \includegraphics[height=2.35cm, trim={0cm 1cm 0cm 0},clip]{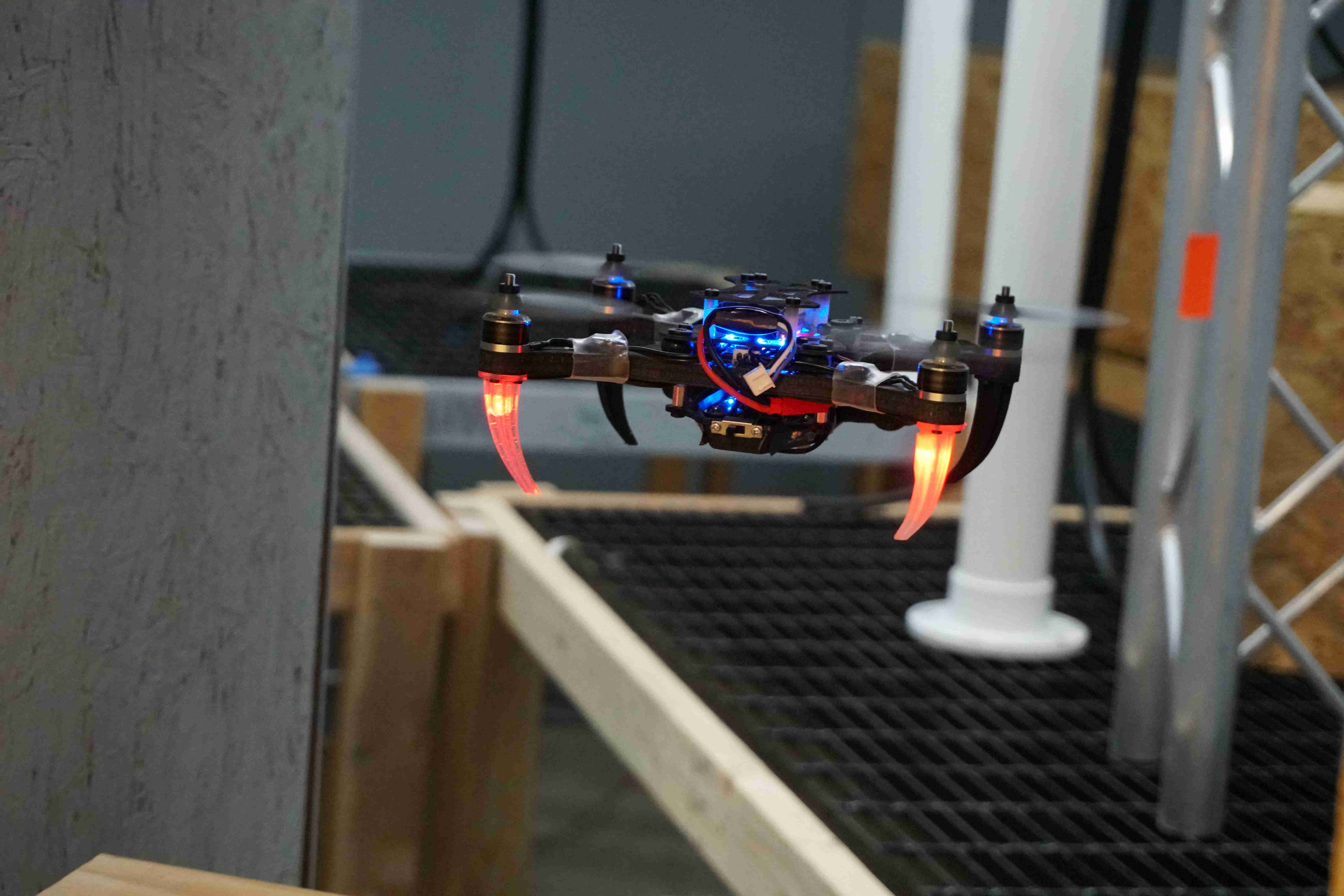}
  \caption{The small 236 g platform used for the nuclear inspection task.}
  \label{fig:platform}
\end{figure*}

The platform is a 0.16 m diameter, 236 g quadrotor (Fig.~\ref{fig:platform}) using $\text{Qualcomm}^{\circledR}$, $\text{Snapdragon}^\text{TM} \text{Flight}^\text{TM}$ with a 2 cell Lipo battery 7.4 V. The proposed solution is based on our previous work~\cite{LoiannoRAL2017}, where we focused on the ability to perform aggressive maneuvers while tracking a simple trajectory. In the proposed scenario, there are several additional challenges that need to be addressed. The vehicle needs to navigate considering different illumination conditions, concurrently creating a map of the environment and replanning its path to avoid obstacles, while reaching the final mission goal. In the following, we present a brief overview of the key onboard approaches for estimation, mapping and planning.

\subsection{State Estimation and Control} 

As shown in~\cite{LoiannoRAL2017}, the Visual Inertial Odometry (VIO) system localizes the rigid body with respect to the inertial frame combining Inertial Measurement Unit (IMU) data and downward facing Video Graphics Array (VGA) resolution camera with $170^\circ$ field of view. The prediction step is based on the IMU integration. The measurement update step is given by the standard $3$D landmark perspective projection onto the image plane leading to the Extended Kalman Filter (EKF) updates. An additional Unscented Kalman Filter (UKF) is able to estimate the state of the vehicle for control purpose at 500 Hz. From a control point of view, we use a proportional and derivate nonlinear controller both for the position and attitude loops~\cite{LoiannoRAL2017,McLamrock,Mellinger2011}.
The control inputs $\tau, \mathbf{M}$ are chosen as
\begin{equation}
\footnotesize
\begin{split}
\mathbf{M}=&-k_R\mathbf{e}_R-k_\Omega \mathbf{e}_\Omega + \Omega\times J\Omega-\mathbf{J}\left(\hat{\Omega}R^\top R_C\Omega_C-R^\top R_C\dot{\Omega}_C\right)\\
\tau=&\left(-k_x\mathbf{e}_x-k_v\mathbf{e}_v+mg\mathbf{e}_3+m\ddot{\mathbf{x}}_d\right)\cdot R\mathbf{e}_3
\label{eq:force_control}
\end{split}
\end{equation}
\\with $\ddot{\mathbf{x}}_d$ the desired acceleration, $k_x$, $k_v$, $k_R$, $k_\Omega$ positive definite terms. The subscript $C$ denotes a commanded value. The quantities $\mathbf{e}_R, \mathbf{e}_\Omega, \mathbf{e}_x, \mathbf{e}_v$ are the orientation, angular rate, and translation errors respectively, defined in~\cite{McLamrock, Mellinger2011} and $R$ the orientation, $\Omega$ the angular rate in the body frame.
The attitude loop runs on the Digital Signal Processor (DSP) available on the board at 1 kHz, whereas the position loop runs on the  Advanced RISC Machines (ARM) processing unit concurrently with the estimation and planning pipelines.

\subsection{Mapping} 

The frontward-facing stereo camera is employed to create a dense map of the environment (see Fig.~\ref{fig:platform}). The stereo camera mapping algorithm uses two rectified images of $640\times 480$ resolution at $15$ Hz with $90$ degrees horizontal field of view to determine the location of obstacles in the environment. To generate a high frame rate map for real-time obstacle avoidance, the rectification process of the two images is split into two separate threads. Both images are concurrently rectified within 3 ms on the ARM processor. A block matcher algorithm produces a disparity map by searching for matches along the epipolar line and provides distance information.
The tradeoff between speed and accuracy can be explored by changing the block matcher filter size. Larger filter window sizes generate less noise in disparity maps at a higher computational cost. The disparity information is then used to generate a 3D pointcloud of the environment from the pinhole camera projection model. We use 3D voxel grids to discretize the space and the 3D points are used as votes to obtain a voxel grid occupancy map for planning.

\subsection{Planning} 

Given a pre-generated global polynomial trajectory and depth map of the environment we use the work in ~\cite{vladyslav_planning} for on-line reactive planning. The generated trajectory deviates from the global path according to encountered obstacles. Quintic B-splines are used to ensure the required smoothness of the trajectory, which are continuous up to the forth derivative of position (snap). Locality of trajectory changes due to changes in the control points, this means that is a change in one control point affects only a few segments in the entire trajectory. Closed-form solutions are available for position at a given time, derivatives with respect to time (velocity, acceleration, jerk, snap) and integrals over squared time derivatives, thus allowing real-time performance.

\begin{figure}[t]
  \centering
  \includegraphics[height=3.5cm]{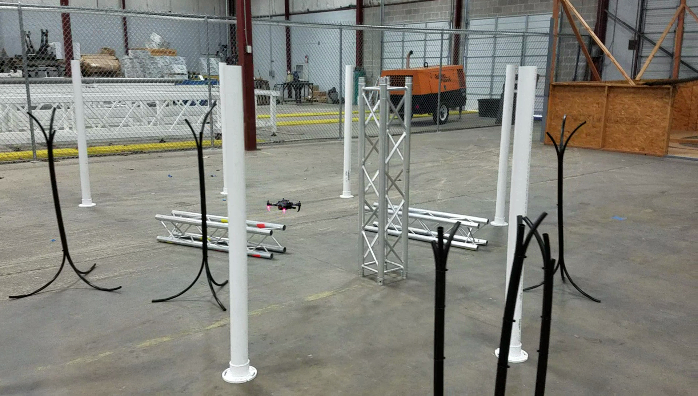}
  \hspace{0.01in}
  \includegraphics[height=3.5cm, trim={9cm 1cm 8cm 9cm},clip]{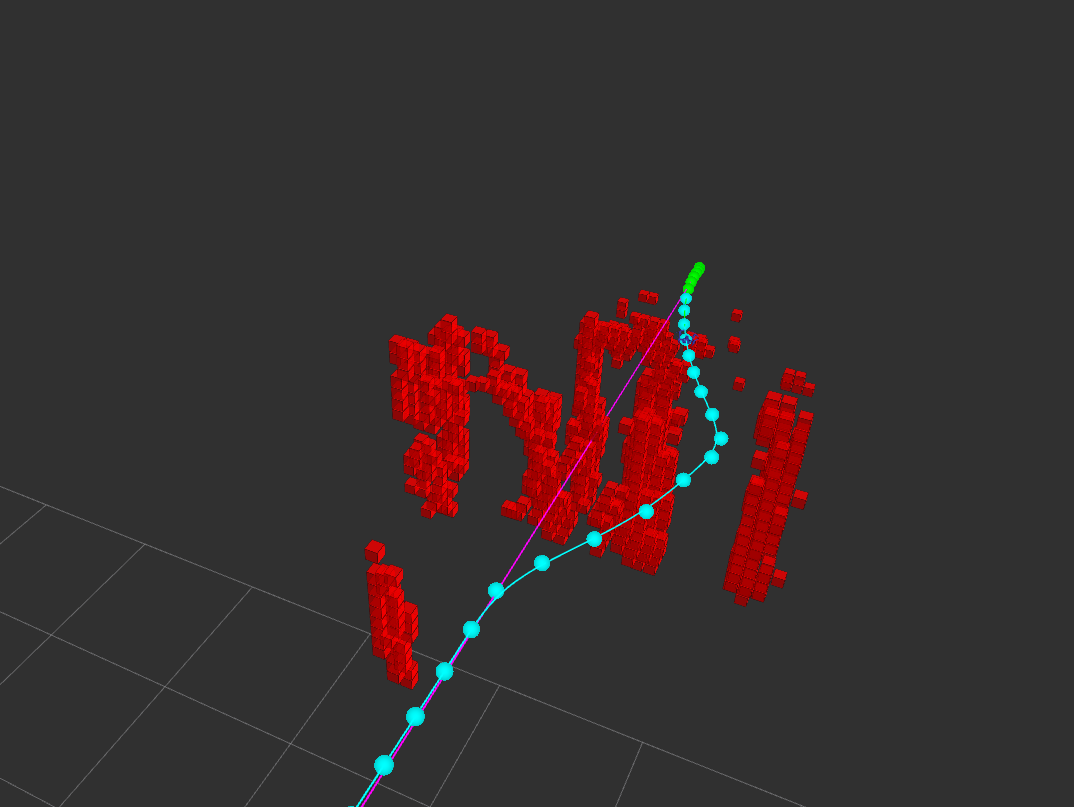}
  \caption{Obstacle course setup (left), Generated map with the path avoiding obstacles.}
  \label{fig:obstacle_course}
\end{figure}

\section{Experimental Results} 
\label{sec:3}

\begin{figure*}[h]
  \centering
  \includegraphics[width=0.45\textwidth]{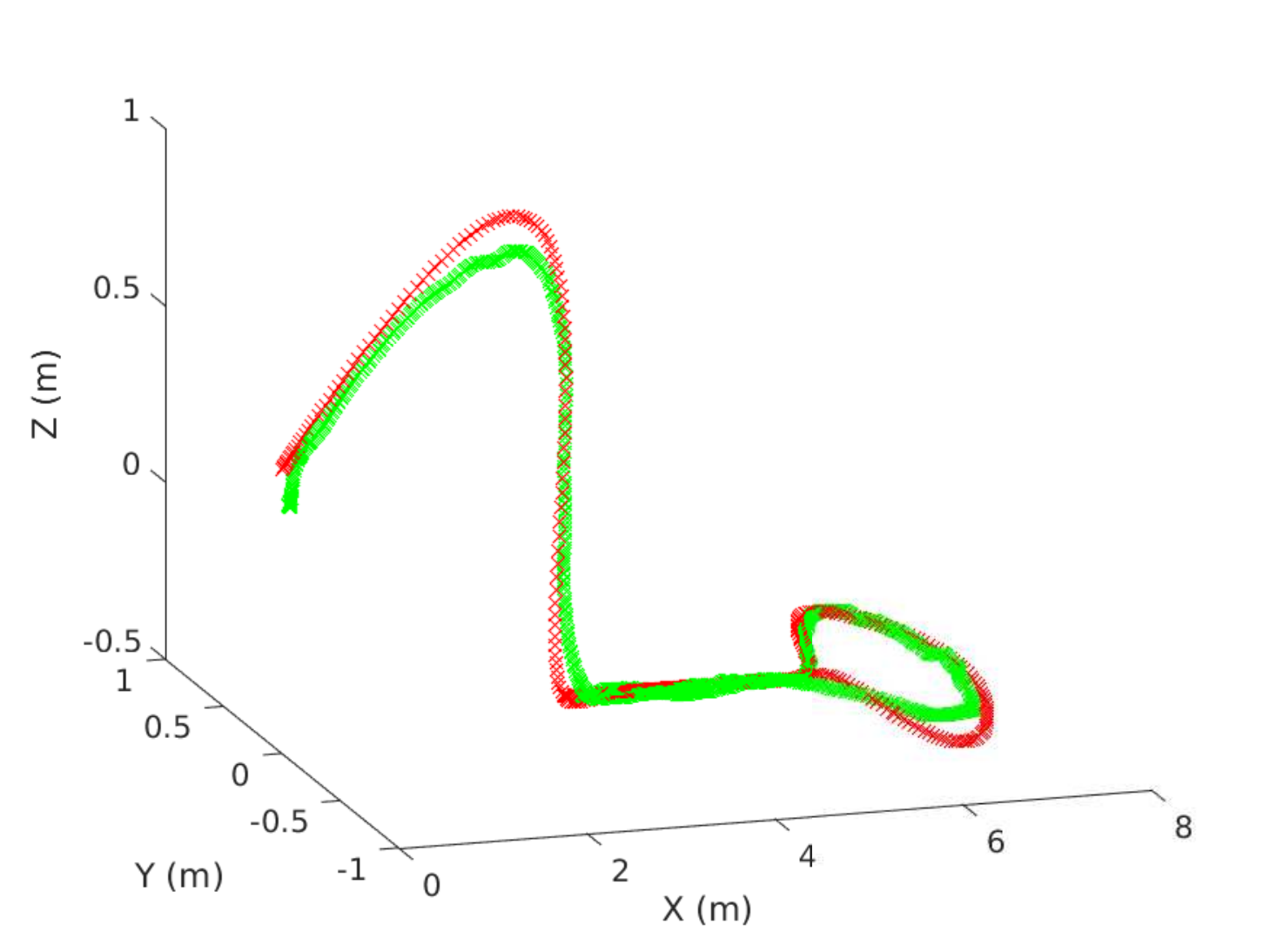}
  \includegraphics[width=0.45\textwidth]{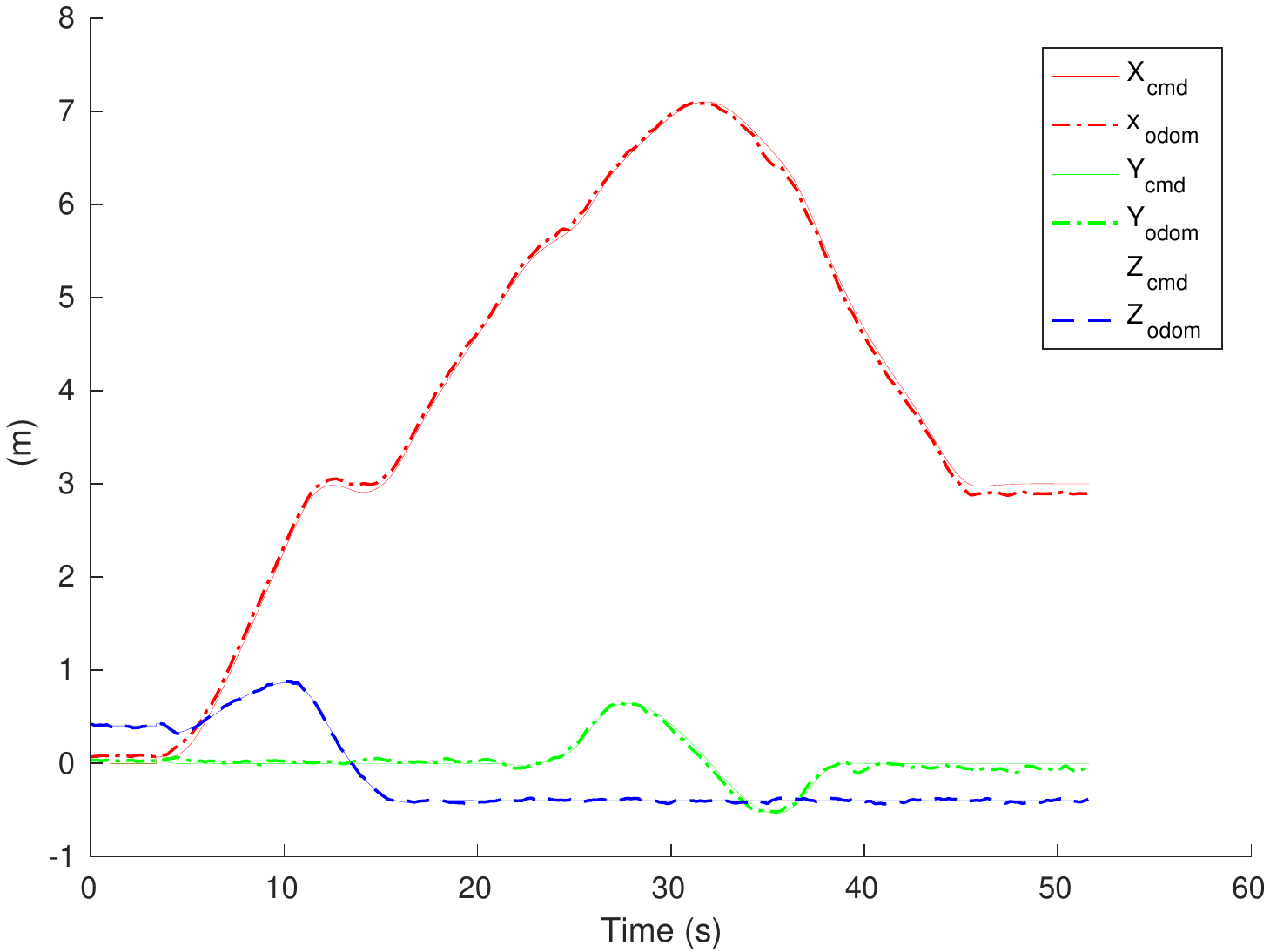}
  \includegraphics[width=0.45\textwidth]{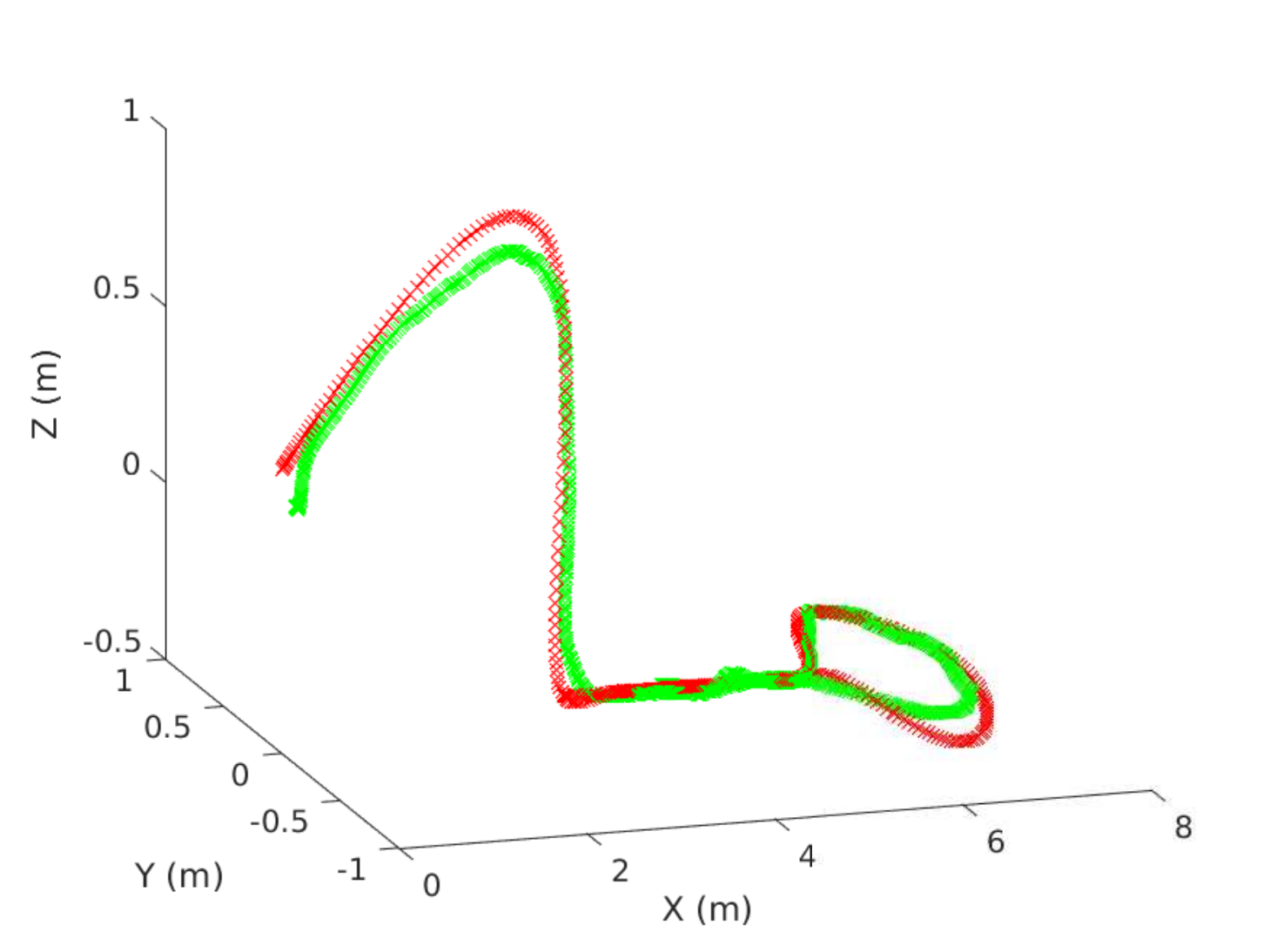}
  \includegraphics[width=0.45\textwidth]{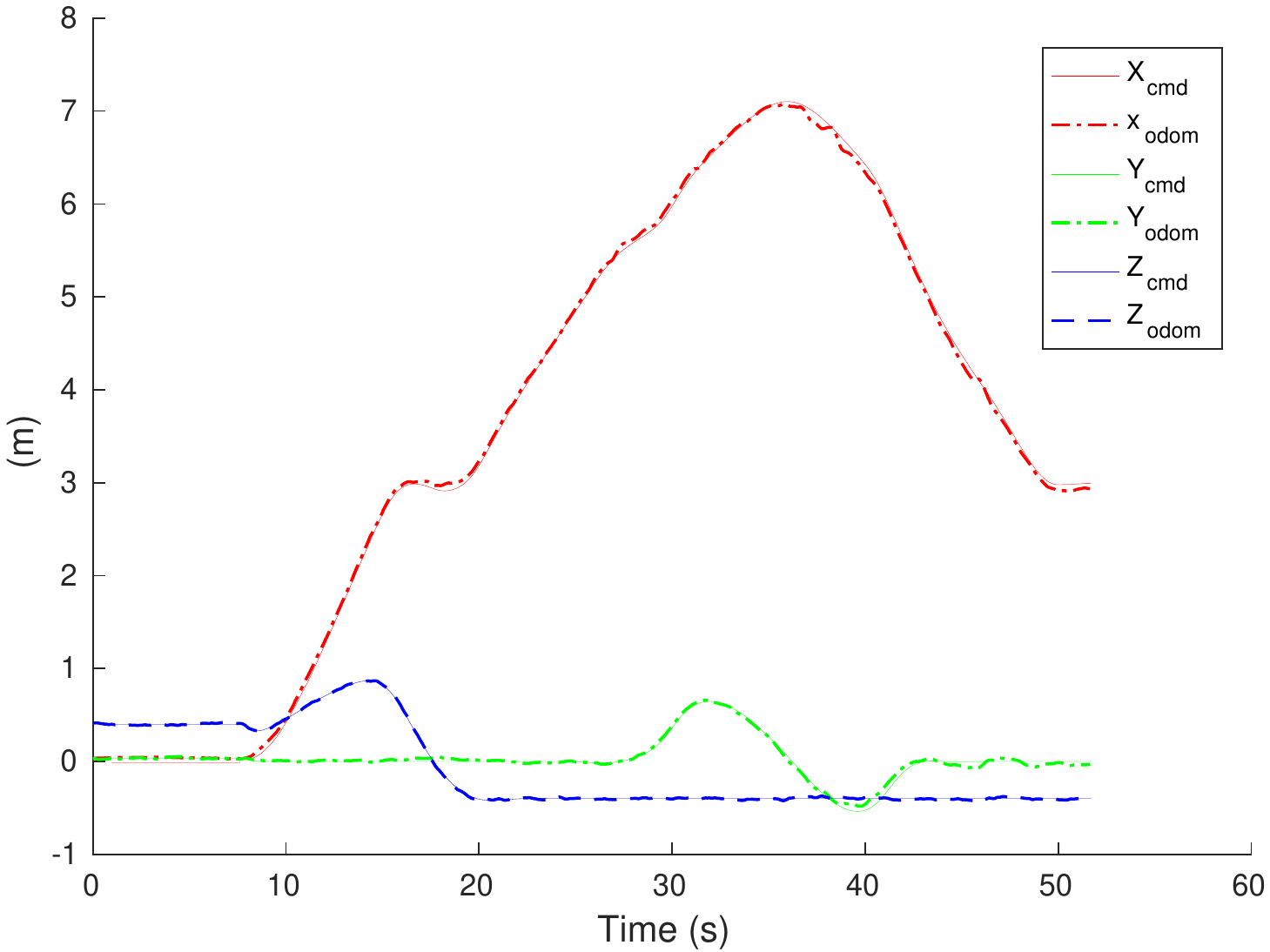}
  \includegraphics[width=0.45\textwidth]{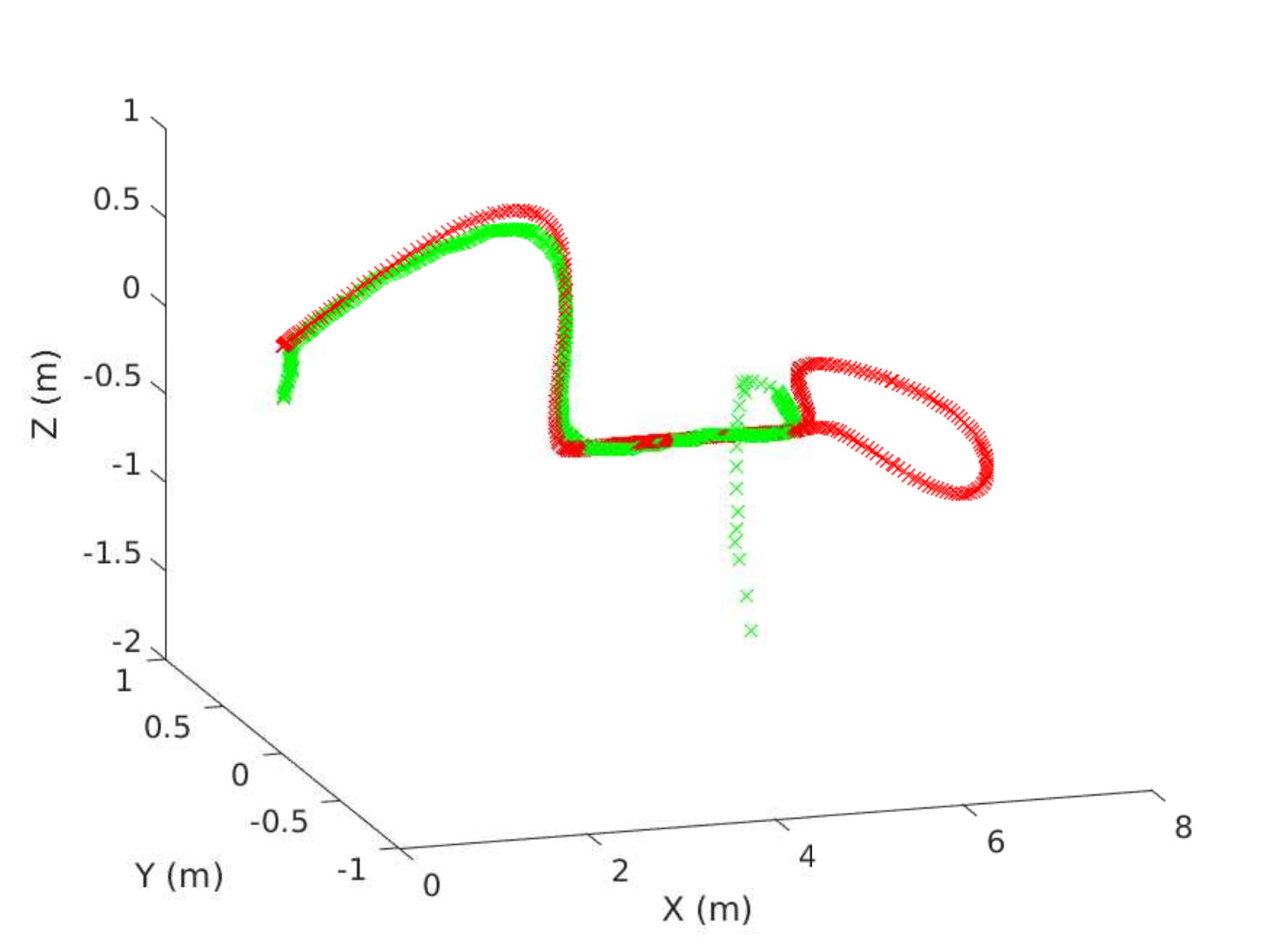}
  \includegraphics[width=0.45\textwidth]{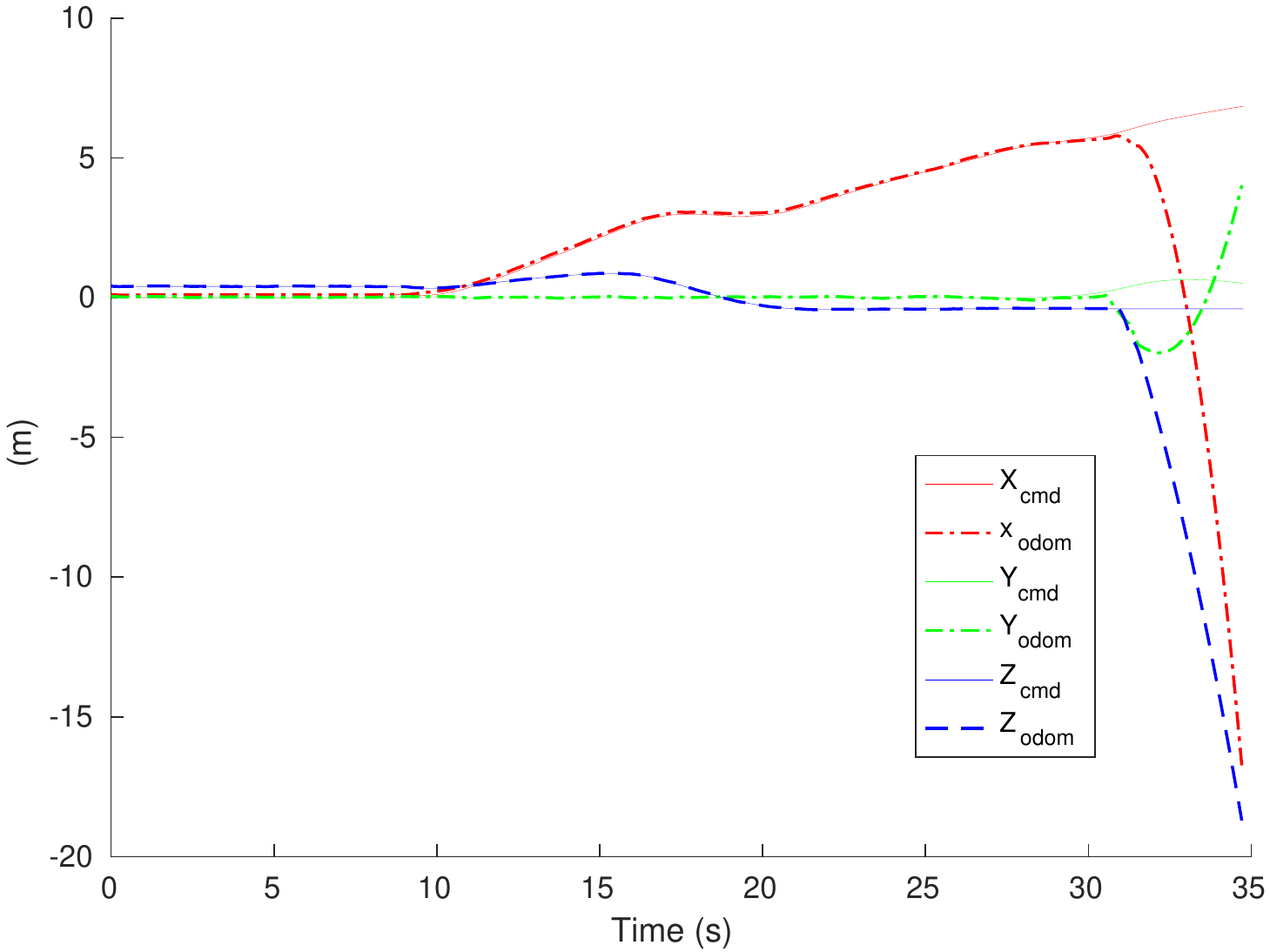}
  \caption{Experiments in different light conditions showing the global trajectory tracking (left column) and tracking error by components (right column) for normal 12 V (up), medium 8 V (center), low 7.5 V (bottom) with crash/drift.}
  \label{fig:pcv_tracking_err}
\end{figure*}

An important step prior to deployment is the evaluation of developed technologies in mock-up facilities. In this section, we report on the experiments carried out in a mock-up of the PCV pedestal (5 m diameter) and CRD ramp as shown in Fig.~\ref{fig:mock_up}. Three different tests were conducted to validate the proposed setup and strategy. During all the tests, the nominal velocity of the vehicle was limited to 0.5m/s.

\subsection{Obstacle avoidance course} 

For these experiments the vehicle is supposed to navigate along an obstacle course that consists of multiple vertical 102 mm PVC pipes, 25.4mm bundled vertical tubes, and aluminum trusses. During the test, the UAS started from a predefined location on the floor and was given a single waypoint located at the far end of the obstacle field. As there were essentially no overhead obstacles, the altitude of the UAS was manually restricted to force it to travel through the obstacle field as opposed to the lower-cost flight path located above the obstacles. Fig.~\ref{fig:obstacle_course} shows the obstacle course and the generated map of one of the multiple obstacle avoidance test trials.

\subsection{Luminance tests} 
\begin{figure*}[t]
  \centering
  \includegraphics[width=0.45\textwidth]{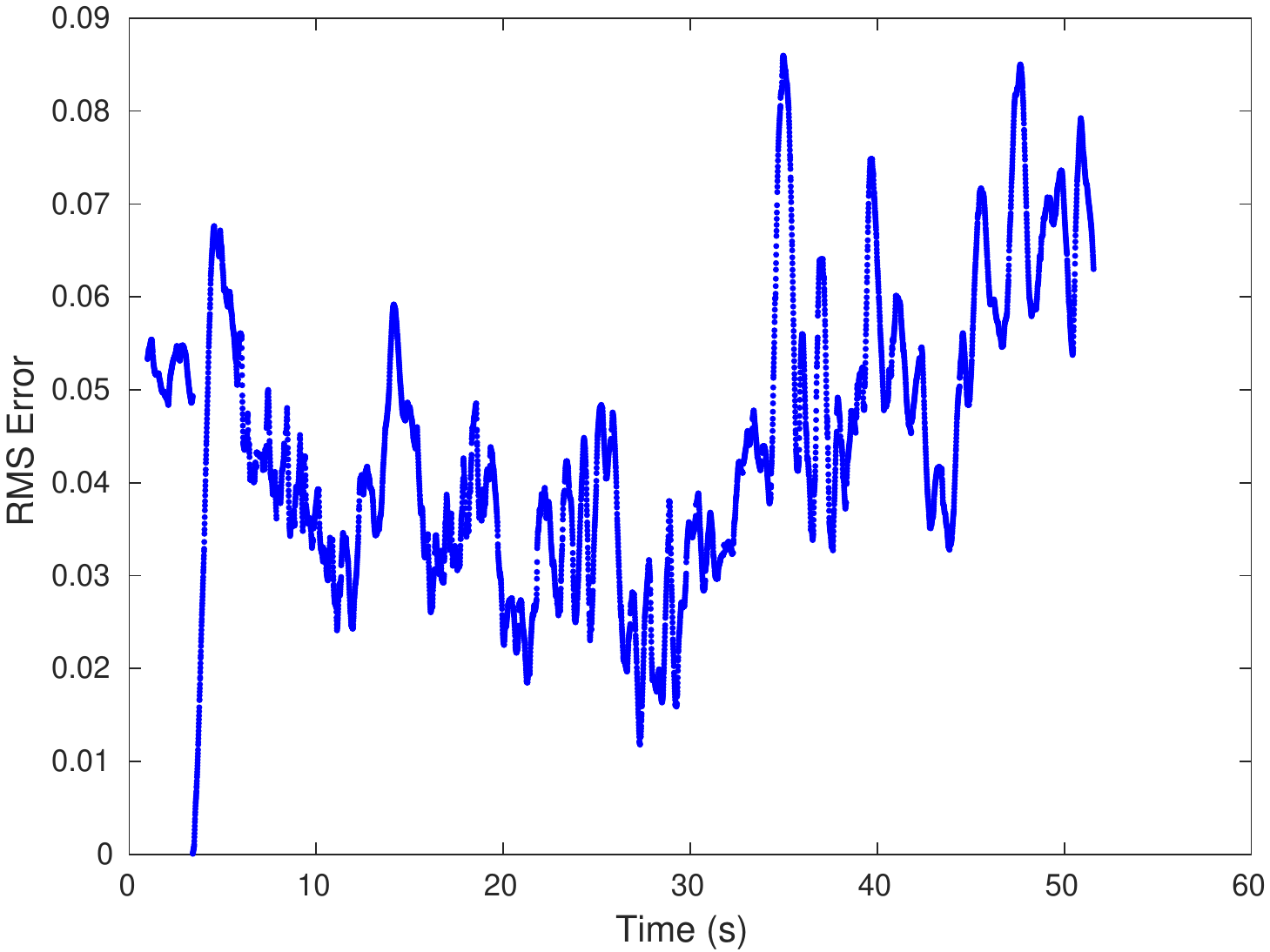}
  \includegraphics[width=0.45\textwidth]{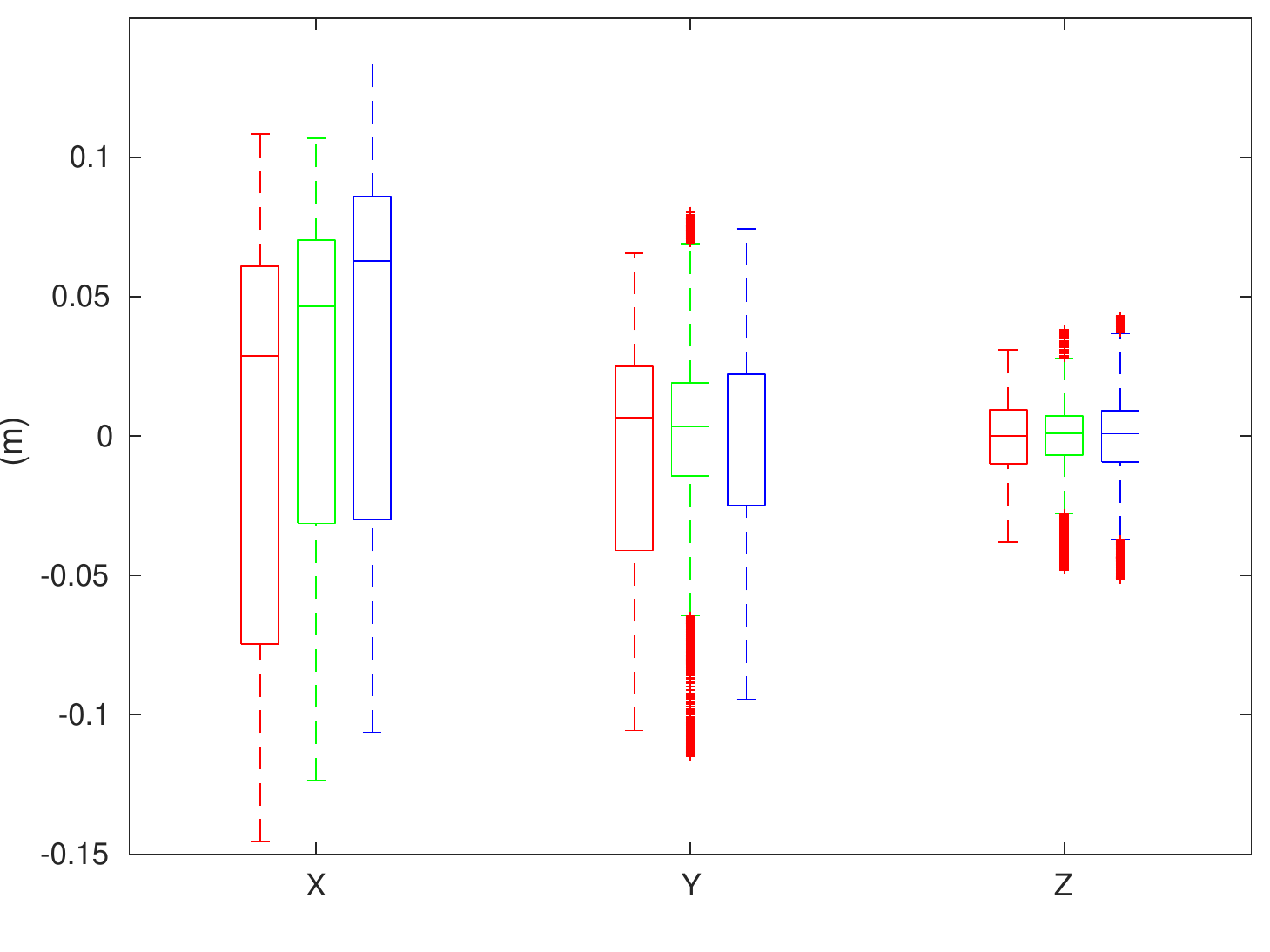}
  \includegraphics[width=0.45\textwidth]{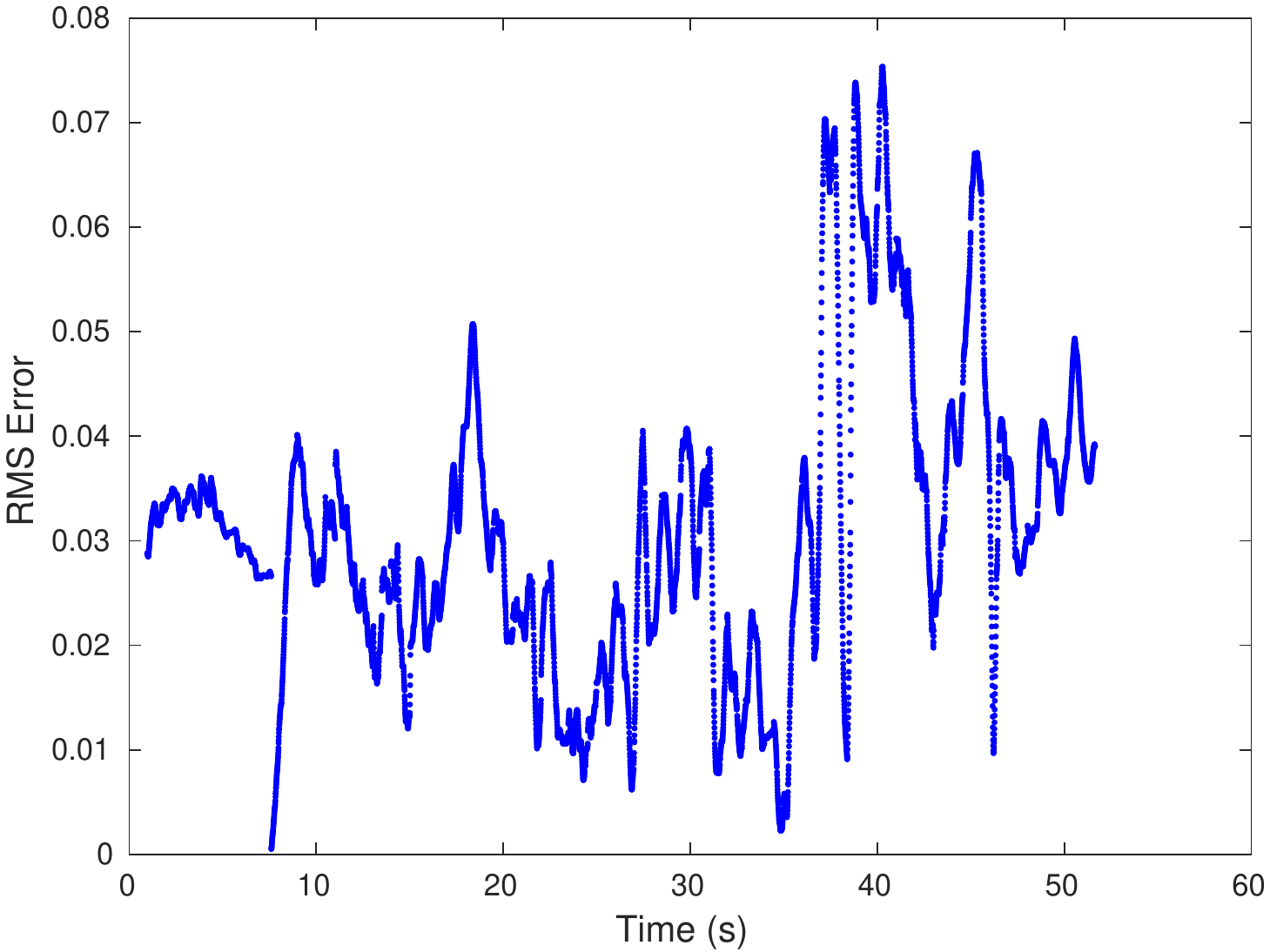}
  \includegraphics[width=0.45\textwidth]{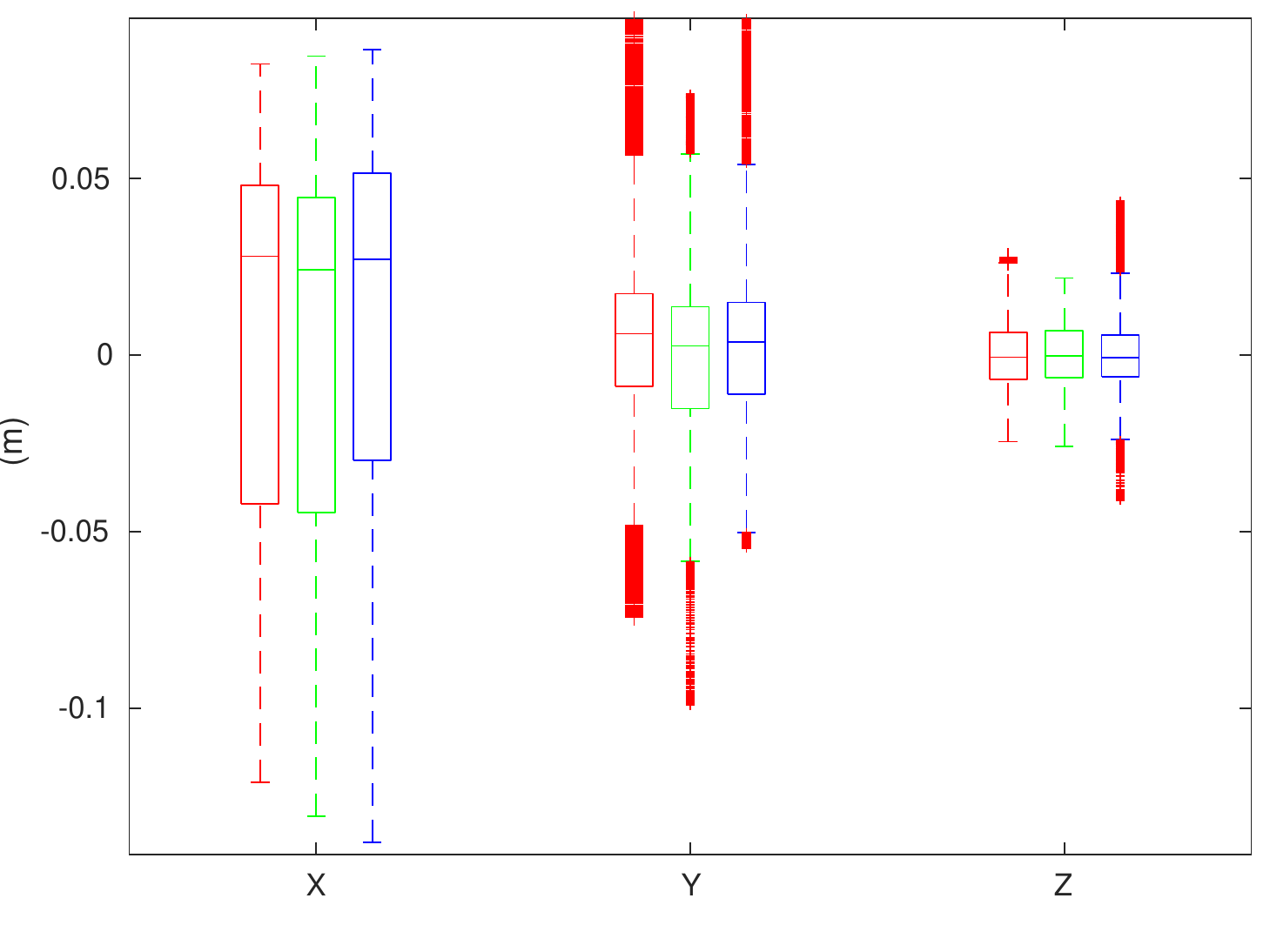}
  \includegraphics[width=0.45\textwidth]{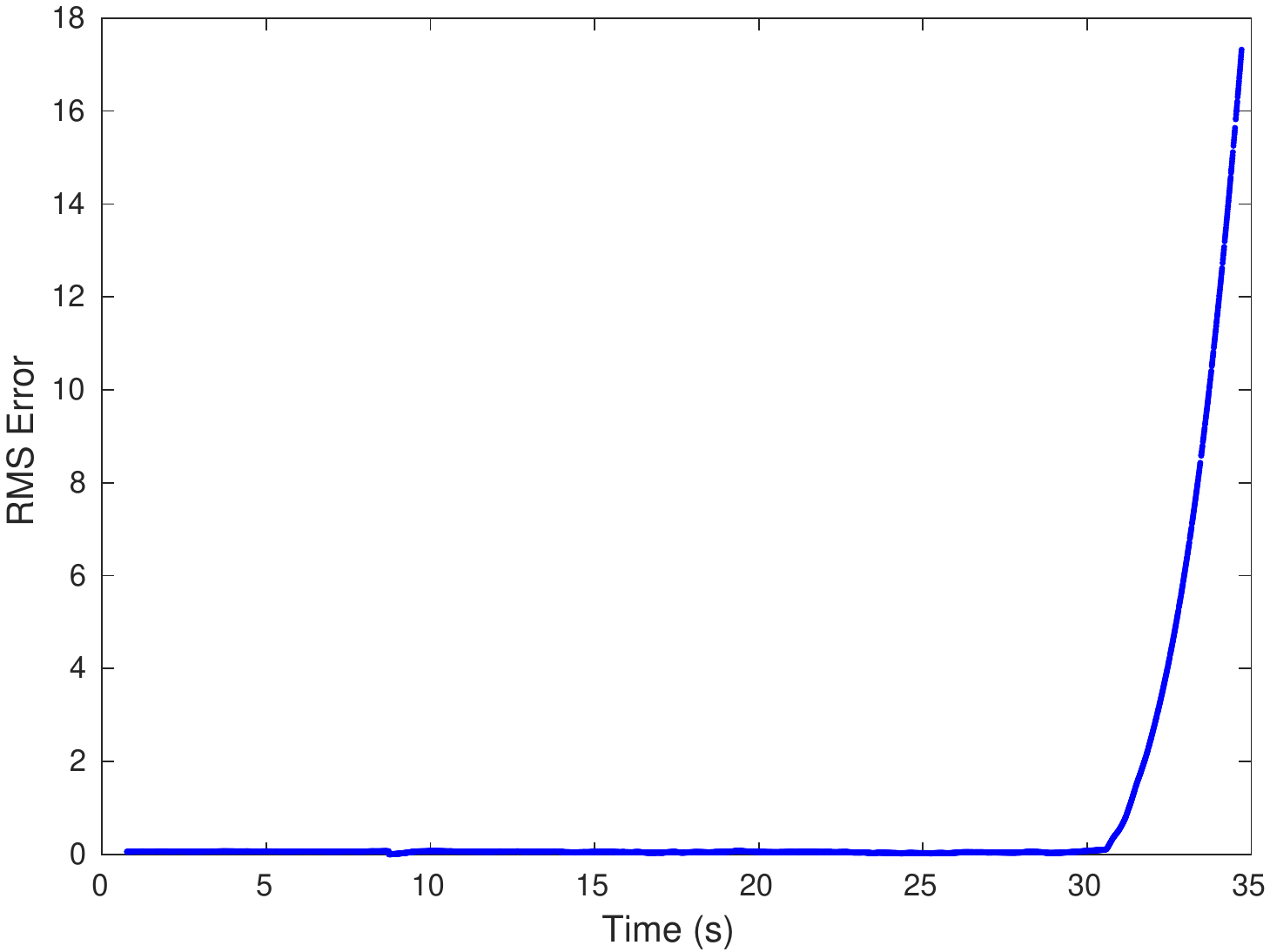}
  \includegraphics[width=0.45\textwidth]{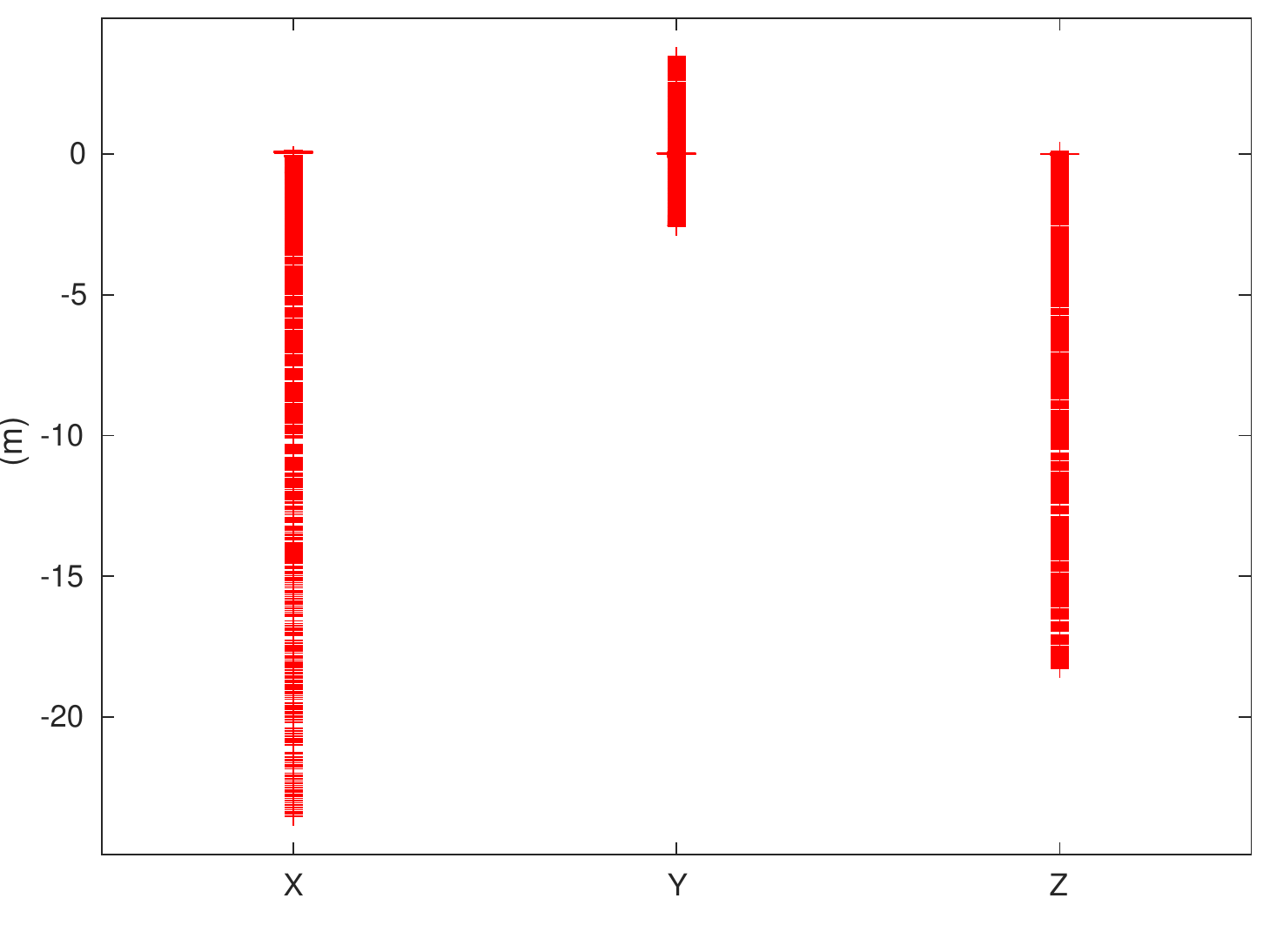}
  \caption{Experiments in different light conditions showing RMSE in X, Y and Z (left column) and maximum errors (right column), for normal 12 V (up), medium 8 V (center), low 7.5 V (bottom) with crash/drift.}
  \label{fig:pcv_rms_err}
\end{figure*}

The current platform is not yet equipped with a lighting source. However, to compensate for variations in light, we can exploit the automatic camera exposure algorithm. To determine an estimate of the illuminance required for the vehicle to accurately extract visual features to maintain adequate localization, experiments with varying luminance were conducted. The PCV pedestal was fitted with multiple LED light strips whose luminous emittance was controlled directly by varying the drive voltage of the lights. Experiments were conducted at three different voltage levels (with all external sources turned off), to find the lowest illuminance to operate the localization algorithm. The corresponding lux values were measured at 7 different locations at constant height above the grates inside the pedestal. Data was collected with an LX1330B digital light meter. The average values corresponding to the voltage levels are 41.8 lx @12V, 9.5 lx @8V, 0.5 lx @7.5V. Fig.~\ref{fig:pcv_tracking_err} (left column) shows the trajectories followed by the UAS in different luminance, whereas in Fig.~\ref{fig:pcv_tracking_err} (right column) we report the control errors between the commanded position and tracked positions. Fig.~\ref{fig:pcv_rms_err} (left column) shows the RMSE (Root Mean Squared Errors), whereas Fig.~\ref{fig:pcv_rms_err} (right column) presents distribution of the tracking error across three runs. Finally, the reader should notice that the 7.5V experiment was performed only once to avoid additional damage to the vehicle due drift the in VIO and subsequent crash.

\subsection{PCV inspection} 
\begin{figure*}[t]
  \centering
  \includegraphics[width=0.9\textwidth]{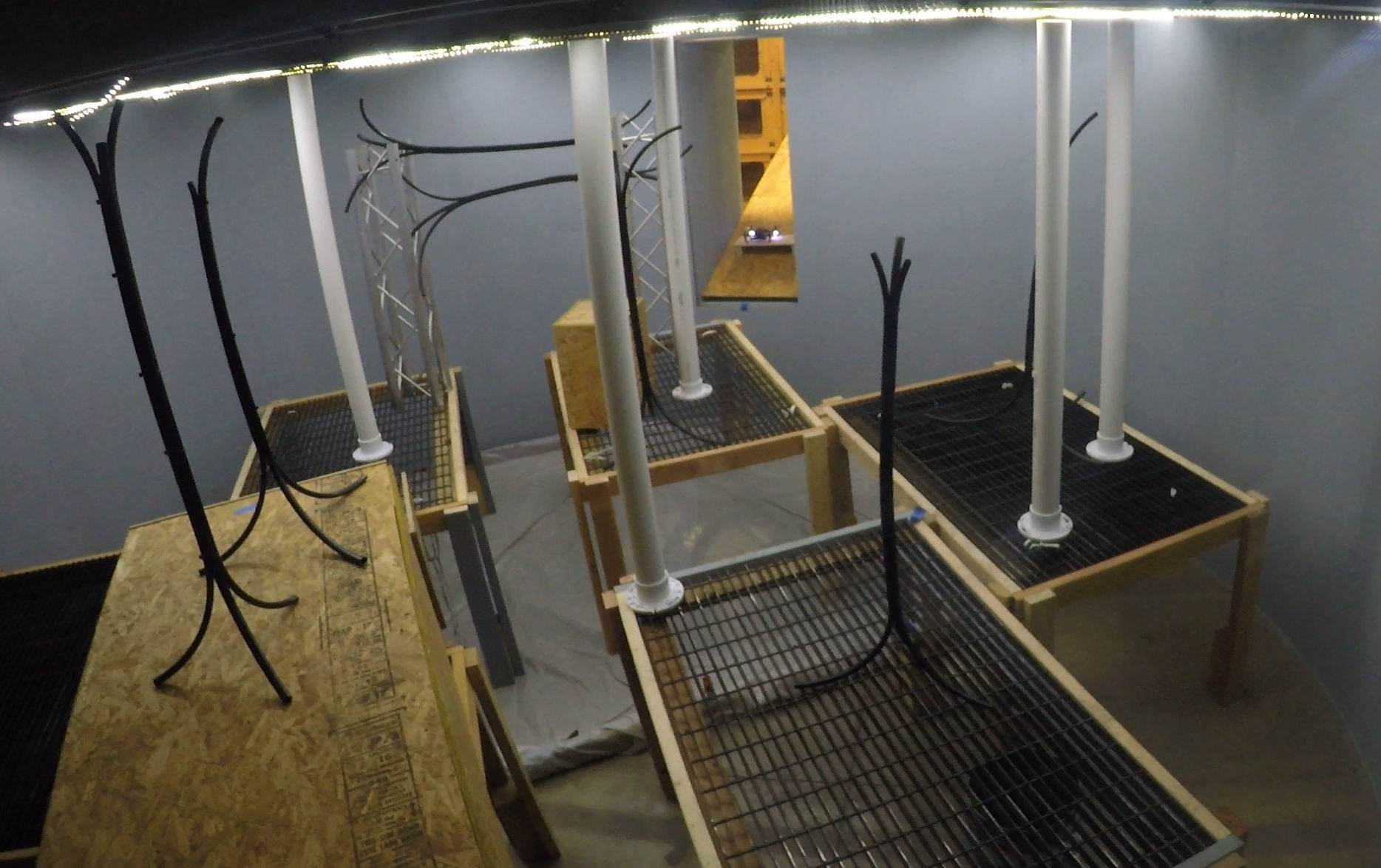}
  \hspace{0.02cm}
  \includegraphics[width=0.45\textwidth]{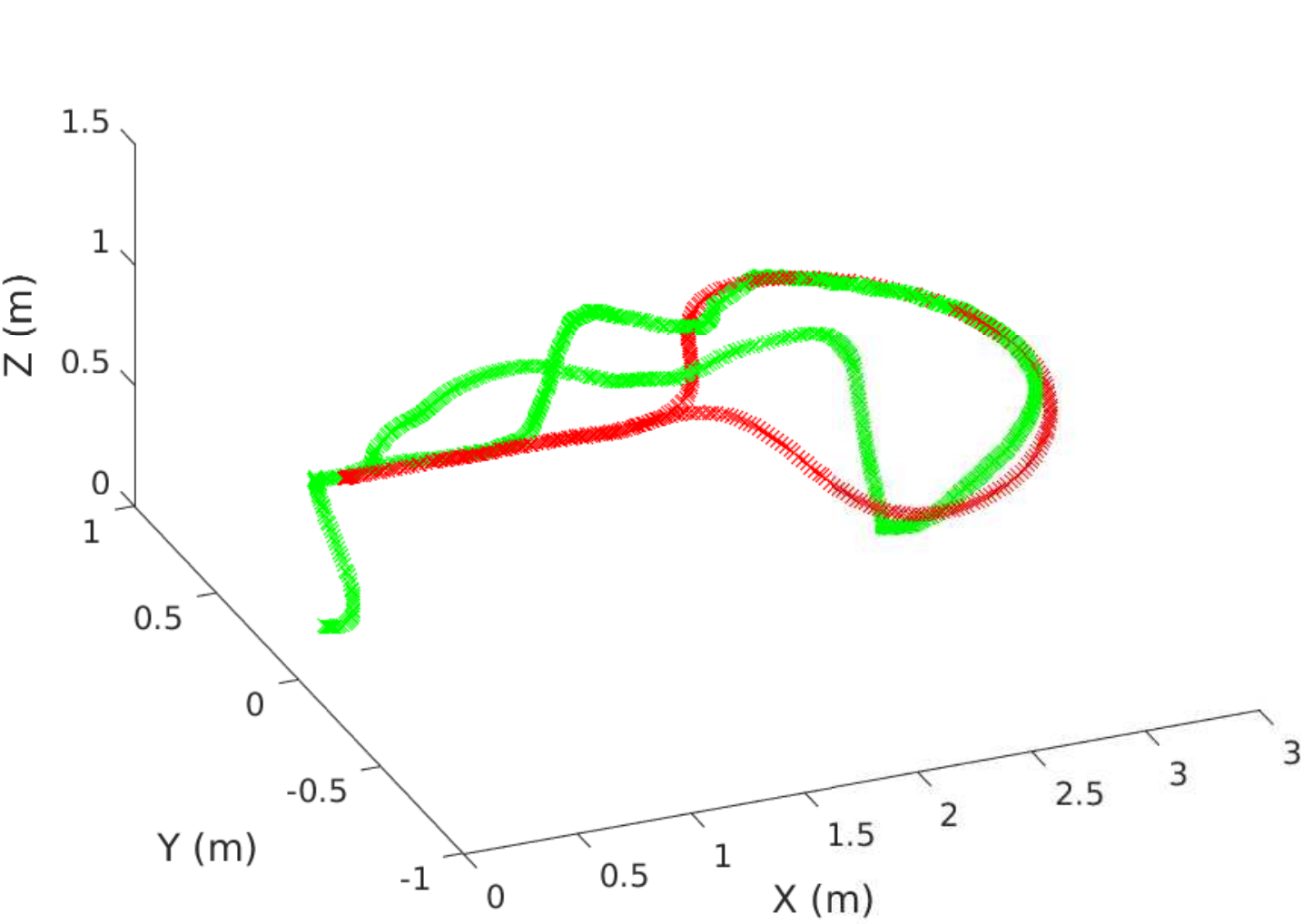}
  \hspace{0.02cm}
  \includegraphics[width=0.45\textwidth]{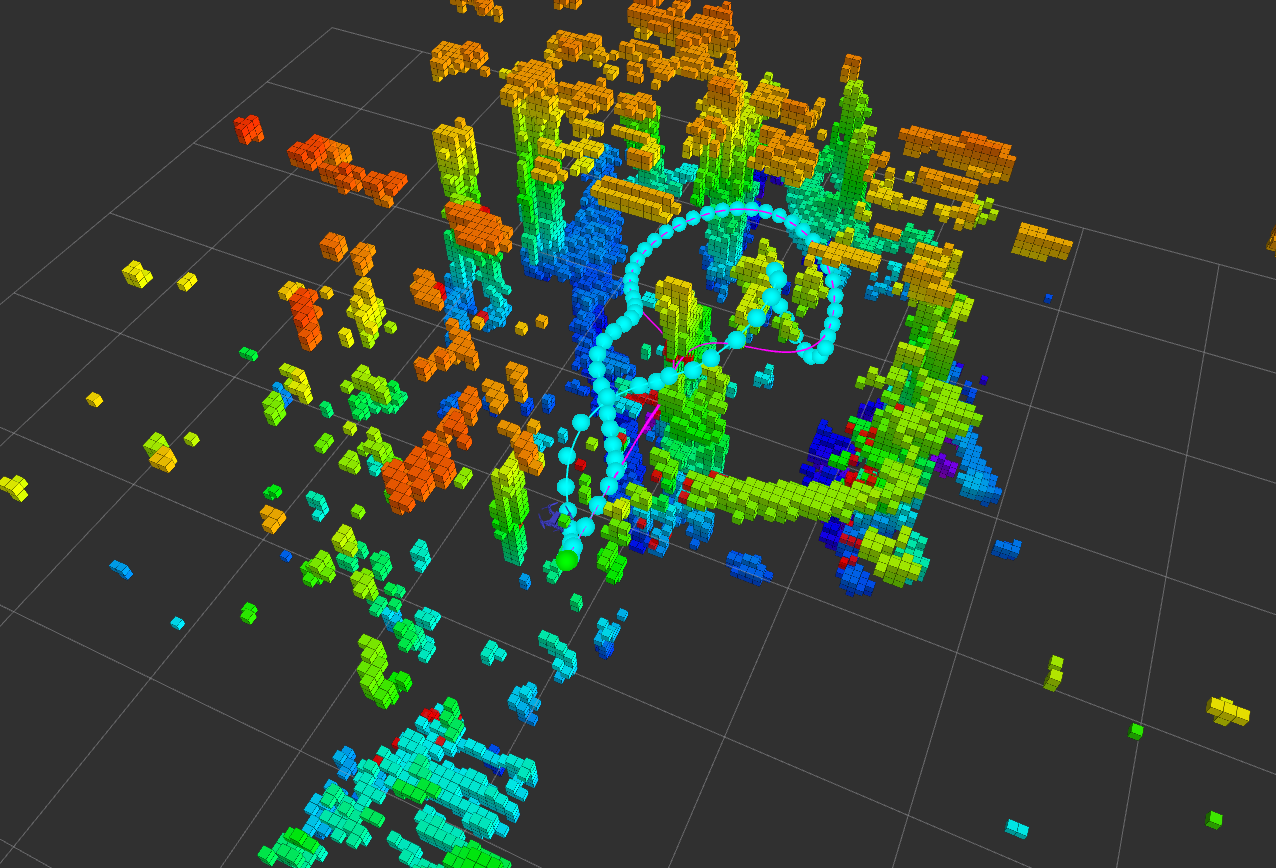}
  \hspace{0.02cm}
  \caption{PCV pedestal setup with unknown obstacles for experiments (top). The red is the original trajectory as if obstacles were not in the path, whereas the green is the state estimate including replanning phases (bottom left). The generated obstacle map (bottom right).}
  \label{fig:pcv_obstacles_exp}
\end{figure*}

The third, and most complex, test scenario consisted of multiple stationary and suspended obstacles (of varying sizes) at unknown locations inside of the pedestal. The vehicle starts from the external point, moves into the cluttered pedestal area and performs the site inspection and environment mapping while avoiding obstacles (Fig.~\ref{fig:pcv_obstacles_exp}). The planning and obstacle avoidance operate in the 3D space and the vehicle safely maneuvers from the pedestal mock-up.
Multiple runs with varying obstacle locations were conducted.

\section{Main Experimental Insights} 
\label{sec:6}

There are several insights that can be obtained from our experiments. First, we showed the possibility to concurrently run, onboard an aerial platform with limited computational capability, the state estimation, control, mapping and planning algorithms with obstacle avoidance to solve a complex nuclear inspection task. Second, we were able to successfully detect and avoid obstacles 0.25m in diameter with a stereo camera even with a relatively small stereo baseline. Finally, experimental evidence shows that an onboard LED payload will be necessary to allow navigation in very dark conditions with illuminance lower than roughly 8 lx.

The replanning strategy can be improved to avoid local minima by utilizing a global planner. To achieve higher speeds, faster sensors and algorithms are needed to locate the obstacles in shorter amount of time. This is essential in operations like the presented one, where there are mission time constraints in addition to spatial ones.
We believe these experimental insights will allow the development of smaller scale platforms up to 0.1 m in diameter and will aid in on-board LED payload design to enable the vehicle to perform autonomous navigation and obstacle avoidance in reduced lighting conditions.

{\bf Acknowledgments.}
We would like to acknowledge the Richard Garcia and Monica Garcia from SwRI who enabled us to conduct these experiments in PCV mock-up at the San Antonio, TX facility. This work was supported by the TEPCO L99048MEC grant, Qualcomm Research, ARL grants W911NF-08-2-0004, W911NF-17-2-0181, ONR grants N00014-07-1-0829, N00014-14-1-0510, ARO grant W911NF-13-1-0350, NSF grants IIS-1426840, IIS-1138847, DARPA grants HR001151626, HR0011516850. This work was supported in part by C-BRIC, one of six centers in JUMP, a Semiconductor Research Corporation (SRC) program sponsored by DARPA.



\end{document}